\documentclass{article}
\usepackage{graphicx}
\graphicspath{{figures/}}
\usepackage[sort&compress, numbers]{natbib}
\usepackage{amsmath}
\usepackage{url}
\usepackage{amssymb}
\usepackage{stmaryrd}
\usepackage{algorithm}
\usepackage{xcolor}
\usepackage{colortbl}
\usepackage{xspace}
\usepackage[footnotesize]{caption}
\usepackage[caption=false,font=footnotesize]{subfig}
\usepackage{tikz}
\usetikzlibrary{arrows}

\usepackage{pgfplots}
\usepackage{multirow}
\usepackage{rotating}

\usepackage{color}
\usepackage{hyperref}

\newcommand{\Nyq}{{\rm nyq}}
\newcommand{\range}[1]{\llbracket #1 \rrbracket}

\newcommand{\st}{{\rm s.t.}}

\DeclareMathOperator*{\argmin}{arg\,min}

\newcommand{\bs}{\boldsymbol}
\newcommand{\bb}{\mathbb}
\newcommand{\cl}{\mathcal}

\newcommand{\ts}{\textstyle}
\newcommand{\ie}{\emph{i.e.},\xspace}
\newcommand{\eg}{\emph{e.g.},\xspace}

\newcommand{\iid}{%
  \ifmmode
  \mathrm{i.i.d.}%
  \else%
  i.i.d.\@\xspace%
  \fi%
}

\usepackage[top=1in, left=1in, right=1in, bottom=1in]{geometry}

\title{Compressive Hyperspectral Imaging:\\ Fourier Transform Interferometry meets Single Pixel Camera}

\author{A. Moshtaghpour$^1$, J. M. Bioucas-Dias$^2$, and L. Jacques$^1$\footnote{The authors thank P. Antoine and M. Roblin (Lambda-X SA, Belgium) for their help in the
acquisition of the FTI measurements.  AM is funded by the FRIA/FNRS. LJ is funded by the F.R.S.-FNRS.}\\
  \footnotesize $^1$ ISPGroup, ICTEAM/ELEN, UCLouvain, Louvain-la-Neuve, Belgium.\\[-1mm]
  \footnotesize $^2$ Instituto de Telecomunica\c c\~oes, Instituto Superior T\'ecnico, Universidade de Lisboa, Portugal.} \date{\empty} 

\renewenvironment{abstract}{\bf {\em\ Abstract---}}{}

\makeatletter
\renewcommand{\section}{\@startsection {section}{1}{\z@}%
             {-3.5ex \@plus -1ex \@minus -.2ex}%
             {2.3ex \@plus.2ex}%
             {\normalfont\large\bfseries}}
\makeatother

\begin{document}

\maketitle
\pagestyle{plain}
\thispagestyle{empty} 
\begin{abstract}
This paper introduces a single-pixel HyperSpectral (HS) imaging framework based on Fourier Transform Interferometry (FTI).  By combining a space-time coding of the light illumination with partial interferometric observations of a collimated light beam (observed by a single pixel), our system benefits from \emph{(i)} reduced measurement rate and light-exposure of the observed object compared to common (Nyquist) FTI imagers, and \emph{(ii)} high spectral resolution as desirable in, \eg Fluorescence Spectroscopy~(FS). From the principles of compressive sensing with multilevel sampling, our method leverages the sparsity ``in level'' of FS data, both in the spectral and the spatial domains. This allows us to optimize the space-time light coding using time-modulated Hadamard patterns. We confirm the effectiveness of our approach by a few numerical experiments.
\end{abstract}
\section{Introduction}
\label{sec:introduction}
\begin{figure}[t]
	\centering
	\includegraphics[width = 0.9\linewidth]{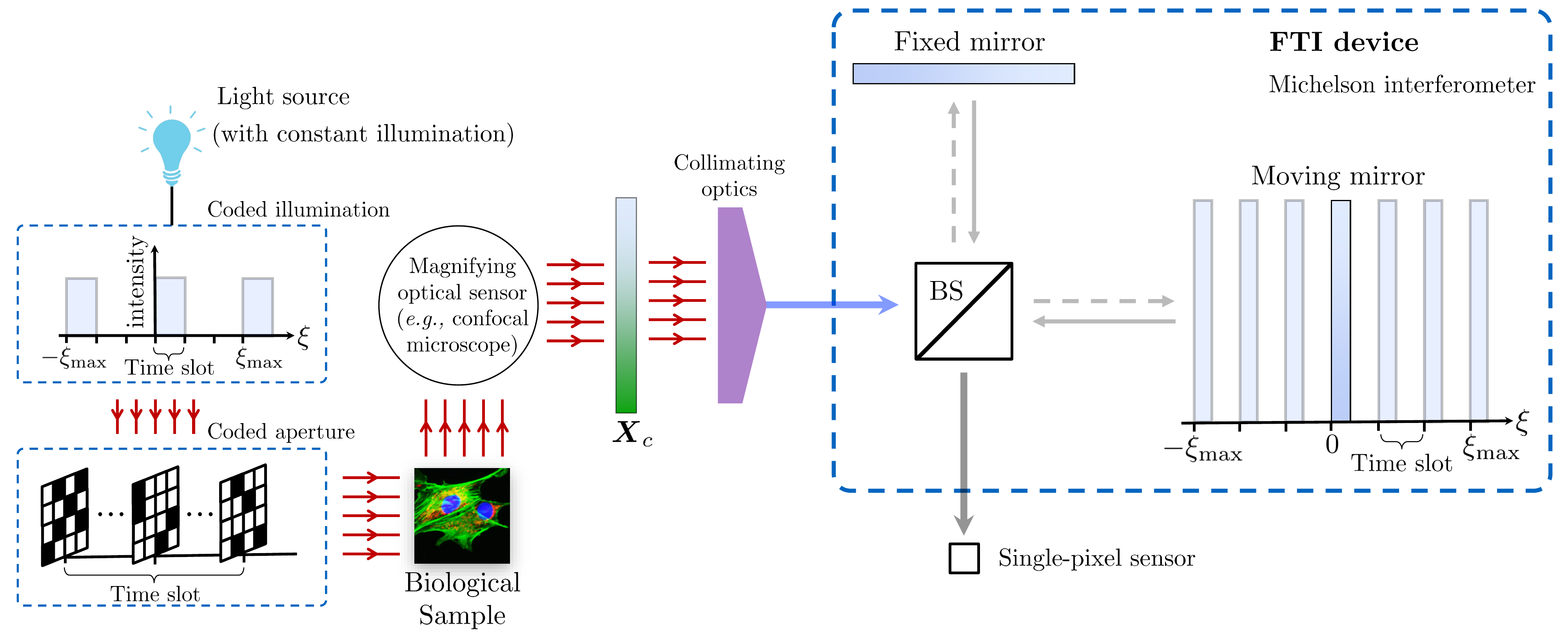}
	\caption{In SP-FTI, a continuous HS volume $\bs X_c$ (\eg observed from a confocal microscope) is spatially and temporally modulated from space-time coding of the light source. On each activated OPD sample, one SP-FTI observation corresponds to the correlation of the HS spatial domain with the CA pattern at this given OPD, as achieved by the Michelson interferometer (dashed box on the right).}
	\label{fig:FTI_scheme}
\end{figure}

Recently, Fourier Transform Interferometry (FTI) has received a renewed interests for capturing HyperSpectral (HS) data where high spectral resolution is desired, \eg in Fluorescence Spectroscopy (FS). As shown in Fig.~\ref{fig:FTI_scheme} (right), FTI works on the principle of a Michelson interferometer with a moving mirror \cite{bell2012introductory}. A coherent wide-band beam entering the FTI device is first divided into two beams by a Beam-Splitter (BS). The resulting beams are then reflected back either by a fixed mirror or by the moving mirror, controlling the Optical Path Difference (OPD) of the two beams, and interfere after being recombined by the BS. The resulting beam, or {\em interferogram}, is later recorded (in intensity) by an external imaging sensor.

Physical optics shows that the outgoing beam from the FTI, as a function of OPD $\xi \in \mathbb{R}$, is the Fourier transform of the entering beam, as a function of wavenumber $\nu \in \mathbb{R}$, \ie $\xi$ and $\nu$ are (Fourier) dual variables. As an advantage, by enlarging the range of recorded OPD values, the spectral resolution of the reconstructed beam (by applying inverse Fourier transform) is increased. On the other hand, in FS applications, this increase of resolution is limited by the durability of the fluorescent dyes when exposed to illumination. Namely, the illumination of fluorescent dyes fades out as they become over-exposed, \ie a phenomenon known as \textit{photo-bleaching} \cite{diaspro2006photobleaching}. 

Several frameworks~\cite{moshtaghpour2016,moshtaghpourcoded,moshtaghpour2017a,moshtaghpour2018,moshtaghpour2018multilevel} have been recently introduced to reduce light exposure in Nyquist FTI -- which senses all OPD samples in a given range -- by leveraging Compressive Sensing (CS) theory \cite{donoho2006compressed,candes2006near}. In \cite{moshtaghpour2016}, a random subsampling of the mirror positions was first considered. Equivalently, \cite{moshtaghpourcoded} replaced this scheme by coding temporally the global light source in a process called Coded Illumination-FTI (CI-FTI). A Coded Aperture-FTI (CA-FTI) was recently proposed in \cite{moshtaghpour2018,moshtaghpour2017a} by modulating spatially the illumination (\eg using a spatial light modulator), hence allowing for different OPD coding per spatial locations, but with no spatial mixing during the acquisition. 

Inspired by successful application of Single-Pixel Camera (SPC)~\cite{duarte2008single} in FS experiments \cite{studer2012compressive,roman2014asymptotic}, this paper proposes a Single-Pixel FTI (SP-FTI) for HS acquisition combining space-time coded illumination (as in CI-FTI and in CA-FTI) with single-pixel acquisition.  
Compared to former CS FTI approaches, SP-FTI allows us to further reduce the light exposure on the observed object by improving the undersampling ratio of the complete system. This is achieved by invoking the recent concept of MultiLevel Sampling (MLS) \cite{adcock2017breaking}. MLS allows us to optimize the space-time light coding with respect to the \emph{sparsity in levels} of FS-HS data, hence boosting HS volumes reconstruction quality in this context.

\section{Proposed Method: SP-FTI}

\noindent\textbf{Acquisition model:} In a simplified setting where $\bs X \in \mathbb{R}^{N_\nu \times N_p}$ is the discretization of $\bs X_c$ (see Fig.~\ref{fig:FTI_scheme}) over $N_\nu = N_\xi$ wavenumber samples and $N_p$ pixels, the SP-FTI measurement matrix $\bs Y \in \mathbb{R}^{M_\xi \times M_p}$ is modeled as
\begin{equation}
\bs Y = \bs P_{\Omega_\xi} \bs F \bs X \bs H^\top \bs P_{\Omega_p}^\top + \bs N,~~~\bs N := \bs P_{\Omega_\xi} \bs N^\Nyq \bs P_{\Omega_p}^\top,\label{eq:SPFTI acquisition model}
\end{equation}
where $\bs P_{\Omega_\xi} \in \{0,1\}^{M_\xi \times N_\xi}$  is the operator extracting the $M_\xi:=|\Omega_\xi|$ rows of a matrix indexed in $\Omega_\xi \subset \range{N_\xi}:=\{1,\cdots,N_\xi\}$, and similarly for $\bs P_{\Omega_p} \in \{0,1\}^{M_p \times N_p}$ with $\Omega_p \subset \range{N_p}$. The matrix $\bs F \in \mathbb{C}^{N_\xi \times N_\xi}$ (resp.~$\bs H \in \{\pm 1/\sqrt{N_p}\}^{N_p \times N_p}$) denotes the 1-D DFT (resp. 2-D Hadamard) matrix, $\bs N^\Nyq = \{n^\Nyq_{l,j}\} \in \mathbb{R}^{N_\xi \times N_p}$ is an additive Gaussian noise with $n^\Nyq_{l,j} \sim_{\iid} \mathcal{N}(0,\sigma_\Nyq^2)$.

In SP-FTI, for a given $N_\xi$ and $N_p$, we first activate the light source during $M_\xi \ll N_\xi$ OPD values, as shown in Fig.~\ref{fig:FTI_scheme} (top-left). In this case, $\Omega_\xi$ corresponds to the indices of active OPDs. At each (active) OPD sample a coded aperture is programed $M_p \ll N_p$ times such that for every aperture pattern only a group of spatial locations of the observed object are exposed. The aperture patterns, at every OPD point, are generated according to the rows of the Hadamard matrix indexed in $\Omega_p$.
A spatially coded HS light beam is then integrated into a single beam, \eg by means of an optical collimator. Following the previous discussion, the FTI outgoing beam is the Fourier transform of the coded and integrated HS light beam.

\noindent \textbf{Advantages:} we consider two criteria, \ie Measurement Undersampling Ratio (MUR) and Exposure Reduction Ratio (ERR). MUR measures the reduction in the total number of measurements, $\textsl{MUR}:= M_\xi M_p/(N_\xi N_p)$, while ERR quantifies the reduction of light exposure on the observed object assuming that the total acquisition time is the same for SP-FTI and Nyquist FTI. Following the structure of the Hadamard matrix, one can compute $\textsl{ERR} = 0.5(1+1/M_p)M_\xi/N_\xi$. Therefore, our HS imaging system can successfully reduce both the number of measurements and the light exposure.  Moreover, regarding the size of the system, as it operates on a collimated beam, SP-FTI requires small-sized mirrors and beam splitter, as opposed to the existing devices. This feature also reduces the effect of optical disturbances, such as diffraction.

\noindent\textbf{HS data recovery:}  as in any other CS applications, HS data recovery requires an accurate low-complexity prior model on the HS volume. Our observations confirm that biological HS data commonly observed in FS share sparse/compressible representation in the Kronecker product of the 1-D Fourier and the 2-D Haar wavelet basis. Accordingly, a stable and robust HS data recovery can be achieved by solving 
\begin{equation}
\ts \hat{\bs X} := \argmin_{\bs U \in \bb R^{N_\xi \times N_p}}  \|\bs \Psi_\nu^\top \bs U \bs \Psi_p\|_1~~~\st~~~\|\bs Y - \bs P_{\Omega_\xi} \bs F \bs U \bs H^\top \bs P_{\Omega_p}^\top\|_F \le \varepsilon, \label{eq:recovery} 
\end{equation}
where $\|\bs V\|_1:=\sum_{l,j} |V_{l,j}|$, $\bs \Psi_\nu \in \mathbb{R}^{N_\xi \times N_\xi}$ (resp. $\bs \Psi_p\in \mathbb{R}^{N_p \times N_p}$) denotes an analysis sparsity basis associated to the spectral (resp. spatial) domain. The parameter $\varepsilon$ must satisfy $\|\bs N \|_F \le \varepsilon$, with $\varepsilon \approx \sigma_\Nyq \sqrt{M_\xi M_p}$ giving satisfactory recovery quality with high probability.

\begin{figure}[t]
	\centering
	 \begin{minipage}{0.49\linewidth}
	 \centering
%
%
\definecolor{mycolor1}{rgb}{0.00000,0.44700,0.74100}%
\begin{tikzpicture}[scale = 1]
\begin{axis}[%
width=2in,
height=1in,
at={(0in,0in)},
scale only axis,
xmin=1,
xmax=32,
xlabel={Sampling level ($t$)},
xtick={1,8,12,16,32},
xticklabels={1,8,12,16,32},
xlabel style={at = {(0.5,0.1)},font=\fontsize{10}{1}\selectfont},
ticklabel style={font=\fontsize{10}{1}\selectfont},
ymin=0,
ymax=1.05,
ytick={0,1},
yticklabels={0,1},
ylabel={$\theta_t$},
ylabel style={at = {(0.18,0.5)},font=\fontsize{10}{1}},
axis background/.style={fill=white},
legend style={at={(0.70,0.93)},anchor=north west,legend cell 
align=left,align=left,draw=none,fill=none,font=\fontsize{8}{1}\selectfont}
]
\node[text=black, draw=none] at (rel axis cs:0.65,0.8) {Spectral};
\node[text=black, draw=none] at (rel axis cs:0.65,0.55) {Fourier/Fourier}; 
\addplot [color=mycolor1,solid,line width=1.0pt,forget plot]
  table[row sep=crcr]{%
1	1\\
2	1\\
3	1\\
4	1\\
5	1\\
6	1\\
7	1\\
8	0\\
9	0\\
10	0\\
11	0\\
12	0\\
13	0\\
14	0\\
15	0\\
16	0\\
17	0\\
18	0\\
19	0\\
20	0\\
21	0\\
22	0\\
23	0\\
24	0\\
25	0\\
26	0\\
27	0\\
28	0\\
29	0\\
30	0\\
31	0\\
32	0\\
};
\end{axis}
\end{tikzpicture}%
	   \end{minipage}
	 \begin{minipage}{0.49\linewidth}
	 \centering
%
%
\definecolor{mycolor1}{rgb}{0.00000,0.44700,0.74100}%
\begin{tikzpicture}[scale = 1]
\begin{axis}[%
width=2in,
height=1in,
at={(0in,0in)},
scale only axis,
xmin=1,
xmax=7,
xlabel={Sampling level ($t$)},
xtick={1,2,3,4,5,6,7},
xticklabels={1,2,3,4,5,6,7},
xlabel style={at = {(0.5,0.1)},font=\fontsize{10}{1}\selectfont},
ticklabel style={font=\fontsize{10}{1}\selectfont},
ymin=0,
ymax=1.05,
ytick={0,1},
yticklabels={0,1},
ylabel={$\theta_t$},
ylabel style={at = {(0.18,0.5)},font=\fontsize{10}{1}},
axis background/.style={fill=white},
legend style={at={(0.47,0.93)},anchor=north west,legend cell 
align=left,align=left,draw=none,fill=none,font=\fontsize{8}{1}\selectfont}
]
\node[text=black, draw=none] at (rel axis cs:0.72,0.8) {Spatial};
\node[text=black, draw=none] at (rel axis cs:0.72,0.55) {Had./Haar}; 
\addplot [color=mycolor1,solid,line width=1.0pt,forget plot]
  table[row sep=crcr]{%
1	1\\
2	1\\
3	0.8125\\
4	0.380208333333333\\
5	0.2421875\\
6	0.187825520833333\\
7	0.0946451822916667\\
};
\end{axis}
\end{tikzpicture}%
	    \end{minipage}
	    \caption{The values of sampling profile per sampling level. For the spectral domain, we suppose symmetric (around the DC frequency) sampling/sparsity levels with identical cardinality. For the spatial domain, we assign sparsity levels to the natural dyadic wavelet levels, also associated with dyadic bands for the sampling levels.}
	    \label{fig:SamplingProfile}
\end{figure}
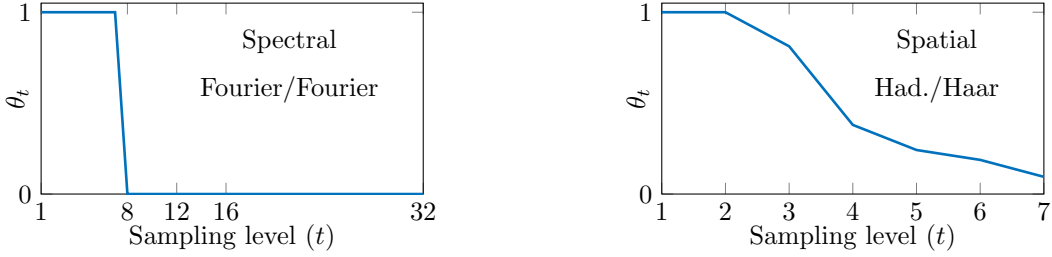

\noindent\textbf{Sampling strategy:} inspired by successful application of MLS \cite{adcock2017breaking} in \cite{moshtaghpour2018multilevel}, our approach leverages the local sparsity structure of HS data in FS. Essentially, for a fixed integer $r$, we define $r$ disjoint sampling levels $\mathcal{W}:=\{\mathcal{W}_1,\cdots,\mathcal{W}_r\}$ (resp. sparsity levels $\mathcal{T}:=\{\mathcal{T}_1,\cdots,\mathcal{T}_r\}$) such that $\bigcup_{t=1}^{r}\mathcal{W}_t = \bigcup_{\ell=1}^{r}\mathcal{T}_\ell =\range{N}$. Roughly speaking, in order to reconstruct a sparse signal that is $k_\ell$-sparse in every $\ell^\textsl{th}$ sparsity level ($ \|\bs P_{\cl T_\ell} \bs \Psi^* \bs x\|_0 \!\le \! k_\ell$) we must reach~\cite{adcock2017breaking} 
$$
\ts m_t = O\big(\,|\mathcal{W}_t|\, (\sum_{\ell=1}^{r} \mu_{t,\ell}(\bs \Phi, \bs \Psi) k_\ell\,) \,\text{polylog}(N)\,\big)
$$
measurements in each sampling level $\cl W_t$, where $\mu_{t,\ell}^2(\bs \Phi,\bs \Psi):= \| \bs P_{\mathcal{W}_t} \bs \Phi \bs \Psi\|_\infty^2 \| \bs P_{\mathcal{W}_t} \bs \Phi \bs \Psi\bs P^\top_{\mathcal{T}_\ell}\|_\infty^2$ is the multilevel coherence between sensing matrix $\bs \Phi$ and sparsity basis $\bs \Psi$. An underlying idea in MLS is to use a sparsity basis that results in a small value of $\theta_t := \min (1, \sum_{\ell=1}^{r} \mu_{t,\ell}(\bs \Phi, \bs \Psi) k_\ell)$ for all $t \in \range{r}$, referred here as \emph{sampling profile}. Computation of $\mu_{t,\ell}$ and $k_\ell$ for the Fourier/Fourier and 2-D Hadamard/Haar systems gives the values of $\theta$ illustrated in Fig.~\ref{fig:SamplingProfile}. Note that the sampling strategy for the spectral and the spatial dimensions can be treated separately. Regarding the estimation of $k_\ell$ for the spectral dimension, we applied the same approach as in \cite{moshtaghpour2018multilevel} on a collection of 24 spectra from Alexa Fluor family \cite{fluorochromes} for finding the worst-case sparsity ratio. For the spatial dimension we used image set BBBC020 (containing 25 FS images) from the Broad Bioimage Benchmark Collection \cite{ljosa2012}. By dividing each image into $128 \times 128$ patches we extracted the worst sparsity ratio among 2000 different patches. The observations in Fig.~\ref{fig:SamplingProfile} suggest a non-uniform sampling strategy as opposed to the (by now classical) Uniform Density Sampling (UDS) strategy~\cite{foucart2013mathematical,candes2006robust,candes2007sparsity}. In Sec.~\ref{sec:numerical-results}, we examine UDS and the sampling strategy demonstrated in Fig.~\ref{fig:SamplingProfile}.
\section{Numerical Results}
\label{sec:numerical-results}
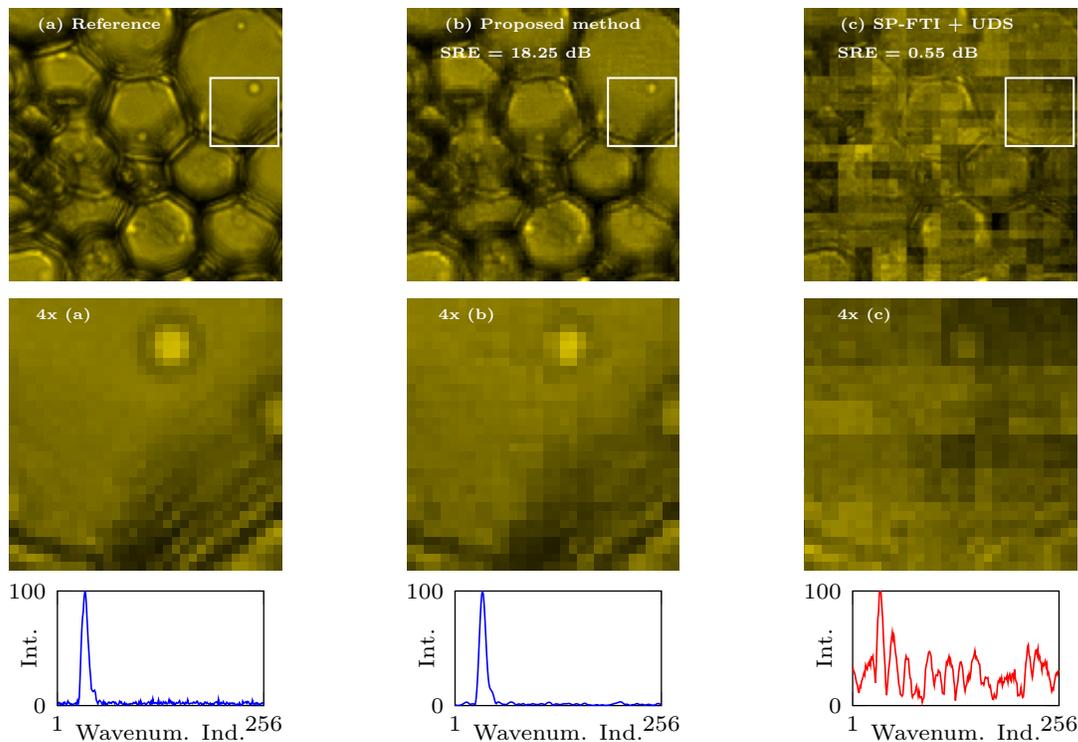
\begin{figure}[tb]
\begin{minipage}{1.01\linewidth}
  \begin{minipage}{0.31\linewidth}
  \centering
%
%
\begin{tikzpicture}[scale = 0.95]

\begin{axis}[%
width=1.5in,
height=1.5in,
at={(0in,0in)},
scale only axis,
axis on top,
xmin=0.5,
xmax=128.5,
y dir=reverse,
ymin=0.5,
ymax=128.5,
hide axis
]
\addplot [forget plot] graphics [xmin=0.5,xmax=128.5,ymin=0.5,ymax=128.5] {Fig_Spatial_ref_DFT-1.png};
\node[fill = none,text=white, draw=none] at (rel axis cs:0.33,0.94) {\fontsize{6}{1}\selectfont \textbf{(a) Reference}}; 
\filldraw[color = white, fill = none, thick] (94.5,32.5)rectangle (126.5,64.5);
\end{axis}
\end{tikzpicture}%
  \end{minipage}
  \begin{minipage}{0.31\linewidth}
  \centering
%
%
\begin{tikzpicture}[scale = 0.95]

\begin{axis}[%
width=1.5in,
height=1.5in,
at={(0in,0in)},
scale only axis,
axis on top,
xmin=0.5,
xmax=128.5,
y dir=reverse,
ymin=0.5,
ymax=128.5,
hide axis
]
\addplot [forget plot] graphics [xmin=0.5,xmax=128.5,ymin=0.5,ymax=128.5] {Fig_Spatial_Dft_Dft-1.png};
\node[fill = none,text=white, draw=none] at (rel axis cs:0.5,0.94) {\fontsize{6}{1}\selectfont \textbf{(b) Proposed method}}; 
\node[fill = none,text=white, draw=none] at (rel axis cs:0.4,0.84) {\fontsize{6}{1}\selectfont \textbf{SRE = 18.25 dB}}; 
\filldraw[color = white, fill = none, thick] (94.5,32.5)rectangle (126.5,64.5);
\end{axis}
\end{tikzpicture}%
  \end{minipage}
  \begin{minipage}{0.31\linewidth}
  \centering
%
%
\begin{tikzpicture}[scale = 0.95]

\begin{axis}[%
width=1.5in,
height=1.5in,
at={(0in,0in)},
scale only axis,
axis on top,
xmin=0.5,
xmax=128.5,
y dir=reverse,
ymin=0.5,
ymax=128.5,
hide axis
]
\addplot [forget plot] graphics [xmin=0.5,xmax=128.5,ymin=0.5,ymax=128.5] {Fig_Spatial_Dft_Dft_Uds-1.png};
\node[fill = none,text=white, draw=none] at (rel axis cs:0.45,0.94) {\fontsize{6}{1}\selectfont \textbf{(c) SP-FTI + UDS}}; 
\node[fill = none,text=white, draw=none] at (rel axis cs:0.38,0.84) {\fontsize{6}{1}\selectfont \textbf{SRE = 0.55 dB}}; 
\filldraw[color = white, fill = none, thick] (94.5,32.5)rectangle (126.5,64.5);
\end{axis}
\end{tikzpicture}%
  \end{minipage}
\end{minipage}
\begin{minipage}{1.01\linewidth}
\vspace{2mm}
  \begin{minipage}{0.31\linewidth}
  \centering
%
%
\begin{tikzpicture}[scale = 0.95]

\begin{axis}[%
width=1.5in,
height=1.5in,
at={(0.729in,0in)},
scale only axis,
axis on top,
xmin=96.5,
xmax=128.5,
y dir=reverse,
ymin=32.5,
ymax=64.5,
hide axis
]
\addplot [forget plot] graphics [xmin=0.5,xmax=128.5,ymin=0.5,ymax=128.5] {Fig_Spatial_ref_DFT-1.png};
\node[fill = none,text=white, draw=none] at (rel axis cs:0.2,0.94) {\fontsize{6}{1}\selectfont \textbf{4x (a)}}; 
\end{axis}
\end{tikzpicture}%
  \end{minipage}
  \begin{minipage}{0.31\linewidth}
  \centering
%
%
\begin{tikzpicture}[scale = 0.95]

\begin{axis}[%
width=1.5in,
height=1.5in,
at={(0.729in,0in)},
scale only axis,
axis on top,
xmin=96.5,
xmax=128.5,
y dir=reverse,
ymin=32.5,
ymax=64.5,
hide axis
]
\addplot [forget plot] graphics [xmin=0.5,xmax=128.5,ymin=0.5,ymax=128.5] {Fig_Spatial_Dft_Dft-1.png};
\node[fill = none,text=white, draw=none] at (rel axis cs:0.22,0.94) {\fontsize{6}{1}\selectfont \textbf{4x (b)}}; 
\end{axis}
\end{tikzpicture}%
  \end{minipage}
  \begin{minipage}{0.31\linewidth}
  \centering
%
%
\begin{tikzpicture}[scale = 0.95]

\begin{axis}[%
width=1.5in,
height=1.5in,
at={(0.729in,0in)},
scale only axis,
axis on top,
xmin=96.5,
xmax=128.5,
y dir=reverse,
ymin=32.5,
ymax=64.5,
hide axis
]
\addplot [forget plot] graphics [xmin=0.5,xmax=128.5,ymin=0.5,ymax=128.5] {Fig_Spatial_Dft_Dft_Uds-1.png};
\node[fill = none,text=white, draw=none] at (rel axis cs:0.22,0.94) {\fontsize{6}{1}\selectfont \textbf{4x (c)}}; 
\end{axis}
\end{tikzpicture}%
  \end{minipage}
\end{minipage}
\begin{minipage}{1.01\linewidth}
 \begin{minipage}{0.31\linewidth}
		  \centering
%
%
\begin{tikzpicture}[scale = 1.2]
\begin{axis}[%
width=0.9in,
height=0.5in,
at={(0in,0in)},
scale only axis,
xmin=1,
xmax=256,
xlabel={Wavenum. Ind.},
xtick={1,256},
xticklabels={1,256},
xlabel style={at = {(0.5,0.35)},font=\fontsize{7}{1}\selectfont},
ticklabel style = {font=\fontsize{7}{1}\selectfont},
ymin=0,
ymax=1,
ytick={0,1},
yticklabels={0,100},
ylabel={Int.},
ylabel style={at = {(0.4,0.5)},font=\fontsize{7}{1}\selectfont},
axis background/.style={fill=white},
legend style={at={(0.53,0.95)},anchor=north west,legend cell align=left,align=left,draw=white!15!black,font=\fontsize{7}{1}\selectfont}
]
\addplot [color=blue,solid,line width=0.5pt,forget plot]
  table[row sep=crcr]{%
1	0.00153183453305501\\
2	0.0209839786505109\\
3	0.0267993824675357\\
4	0.0109279177234657\\
5	0.0207504433900009\\
6	0.0110282693883301\\
7	0.00690008686059249\\
8	0.0230041435176962\\
9	0.0137103010725273\\
10	0.0299883902508113\\
11	0.0210925600018449\\
12	0.00420636517439632\\
13	0.0169849384854894\\
14	0.0391930763198558\\
15	0.0240688901352632\\
16	0.0230405310355335\\
17	0.00615965663998461\\
18	0.0147359791766515\\
19	0.0132104673416007\\
20	0.00716393888297379\\
21	0.0128027936940194\\
22	0.0145474074202127\\
23	0.0207181113170189\\
24	0.0240860928242674\\
25	0.0069182452610713\\
26	0.0457733232249509\\
27	0.0228456419412547\\
28	0.0638948938924645\\
29	0.235324593006825\\
30	0.447485094051885\\
31	0.639042080517455\\
32	0.763458073612088\\
33	0.820636281221386\\
34	0.94248819663112\\
35	1\\
36	0.97479536032221\\
37	0.84799022064868\\
38	0.710115178524258\\
39	0.568785776217646\\
40	0.460516170759936\\
41	0.333251666413104\\
42	0.255389222826303\\
43	0.145531644541575\\
44	0.132742354613454\\
45	0.122062844166718\\
46	0.124531107450872\\
47	0.133311827938419\\
48	0.114662922886886\\
49	0.0516527245642251\\
50	0.0333244569144531\\
51	0.0157421272358528\\
52	0.0343014676753422\\
53	0.0450422441708757\\
54	0.0396857146427494\\
55	0.0312035568255906\\
56	0.0148075783869349\\
57	0.0103591211833702\\
58	0.0355472784739648\\
59	0.0129678822115184\\
60	0.0267817797546852\\
61	0.0115801912098588\\
62	0.0141905560803194\\
63	0.0126748271851735\\
64	0.0287794101472212\\
65	0.0300900975969787\\
66	0.0202124184668069\\
67	0.0253403140629954\\
68	0.0198924857693666\\
69	0.0329569884841655\\
70	0.0338969594349871\\
71	0.0326700882697447\\
72	0.0168811821114902\\
73	0.0264299844777608\\
74	0.0165158158843948\\
75	0.00579871514340775\\
76	0.000910613754425571\\
77	0.0174783165531008\\
78	0.0366539618110962\\
79	0.0245539236427601\\
80	0.0318520701777187\\
81	0.0160711346209581\\
82	0.0120213984983625\\
83	0.00515101511436991\\
84	0.0247990317417898\\
85	0.0125306680239027\\
86	0.017908724526407\\
87	0.00491044216251473\\
88	0.0116309233292565\\
89	0.0130136774261912\\
90	0.0143163431273729\\
91	0.0113231653845022\\
92	0.0170056399416846\\
93	0.016120428299183\\
94	0.0146145276285884\\
95	0.026784103848547\\
96	0.020190701883109\\
97	0.0220558074237072\\
98	0.0320279368762555\\
99	0.0137347215373778\\
100	0.0108885876588402\\
101	0.00950774594778188\\
102	0.034606393204529\\
103	0.016615873281925\\
104	0.0414483534041507\\
105	0.0100748307897226\\
106	0.0174650306810163\\
107	0.0169083872736657\\
108	0.0208961883704288\\
109	0.021727374537642\\
110	0.00774609884520246\\
111	0.00882395916404917\\
112	0.02063488296106\\
113	0.0277939148161255\\
114	0.033220457812358\\
115	0.0235722285183691\\
116	0.00745402448972489\\
117	0.0372789348200425\\
118	0.0175759588524408\\
119	0.0583575514153818\\
120	0.0215782290844395\\
121	0.0182830231226473\\
122	0.0237014113316104\\
123	0.0261489166223899\\
124	0.0184507066054025\\
125	0.0130705040729417\\
126	0.0488930648264428\\
127	0.0147417091476902\\
128	0.0158595149640624\\
129	0.0300295602332523\\
130	0.0181384481542239\\
131	0.00594316516864167\\
132	0.0341225990700048\\
133	0.0287579233199676\\
134	0.0187286986160073\\
135	0.0302928084328646\\
136	0.0156630434508482\\
137	0.0155448739020224\\
138	0.0228447670717752\\
139	0.0513751308392689\\
140	0.0249270830834931\\
141	0.0397945739394752\\
142	0.0212324293493137\\
143	0.0285113231853656\\
144	0.00984153022495184\\
145	0.024667808401369\\
146	0.00783464986968221\\
147	0.0130602434436392\\
148	0.0253670974255308\\
149	0.0198156155603116\\
150	0.0217078171888864\\
151	0.0300208492928006\\
152	0.0413854912456635\\
153	0.0136977403812847\\
154	0.0159601942289171\\
155	0.0189544905970284\\
156	0.0290428857933069\\
157	0.0222716879168134\\
158	0.0115016483348349\\
159	0.0259172448349362\\
160	0.0162498801193879\\
161	0.022467005561008\\
162	0.0508221496861964\\
163	0.0176779820088422\\
164	0.0135966770507528\\
165	0.0178748074653338\\
166	0.0217193758397481\\
167	0.0189302802094731\\
168	0.0168701023434865\\
169	0.0300745158136065\\
170	0.0287014046558909\\
171	0.0202284784154834\\
172	0.0225525551133554\\
173	0.0356777928689399\\
174	0.0383639215320504\\
175	0.0185486782050438\\
176	0.00388314552807096\\
177	0.0206999299080947\\
178	0.0236049458130498\\
179	0.0175279940507014\\
180	0.00203554491484225\\
181	0.0144045617990937\\
182	0.0108570437513083\\
183	0.0291438252976155\\
184	0.0307185308744095\\
185	0.0179402202897627\\
186	0.0269519293196689\\
187	0.0186884573331875\\
188	0.0132946377482813\\
189	0.0138939914100902\\
190	0.0208602286084643\\
191	0.00616858465671181\\
192	0.0147193833136188\\
193	0.0174354212490202\\
194	0.00583724551207479\\
195	0.021145208726144\\
196	0.02645537920865\\
197	0.0321210352726644\\
198	0.0310962948337192\\
199	0.0165422988849577\\
200	0.055741196051891\\
201	0.00142128604053348\\
202	0.0227616476926286\\
203	0.0278680044235774\\
204	0.0148600000422148\\
205	0.019063023518721\\
206	0.0336301826704831\\
207	0.0166540590380662\\
208	0.0152041805385674\\
209	0.00257568903146536\\
210	0.028524976655735\\
211	0.0260728532296324\\
212	0.0151915475995326\\
213	0.00437965482998258\\
214	0.0215667313998453\\
215	0.0159724150224621\\
216	0.0210609584013732\\
217	0.0229655607213924\\
218	0.0141234569917067\\
219	0.0216175598446246\\
220	0.0177854811536605\\
221	0.00206112054134618\\
222	0.00387336641110527\\
223	0.0134207452125757\\
224	0.0061843500680438\\
225	0.0114939546386827\\
226	0.025587442353299\\
227	0.00977231792831841\\
228	0.00427311284500303\\
229	0.00416134523485299\\
230	0.0365412922359959\\
231	0.0182397959483919\\
232	0.0271592363788607\\
233	0.0283207100076731\\
234	0.0324125923413497\\
235	0.0193076503859473\\
236	0.0359888373149785\\
237	0.0340846543710427\\
238	0.0153030571231611\\
239	0.0113711967890085\\
240	0.0188635570040937\\
241	0.0111121439636132\\
242	0.0463148493632899\\
243	0.0209674442766485\\
244	0.00810533517193643\\
245	0.0156568734985293\\
246	0.0228291175051871\\
247	0.01206208538312\\
248	0.0238231556067343\\
249	0.0176865779651936\\
250	0.00250201622664749\\
251	0.0131600786005192\\
252	0.00906894730377064\\
253	0.0177380364076697\\
254	0.024193731000842\\
255	0.0199024588061166\\
256	0.00616724636999055\\
};
\end{axis}
\end{tikzpicture}%
\end{minipage}
 \begin{minipage}{0.31\linewidth}
		  \centering
%
%
\begin{tikzpicture}[scale = 1.2]
\begin{axis}[%
width=0.9in,
height=0.5in,
at={(0in,0in)},
scale only axis,
xmin=1,
xmax=256,
xlabel={Wavenum. Ind.},
xtick={1,256},
xticklabels={1,256},
xlabel style={at = {(0.5,0.35)},font=\fontsize{7}{1}\selectfont},
ticklabel style = {font=\fontsize{7}{1}\selectfont},
ymin=0,
ymax=1,
ytick={0,1},
yticklabels={0,100},
ylabel={Int.},
ylabel style={at = {(0.4,0.5)},font=\fontsize{7}{1}\selectfont},
axis background/.style={fill=white},
legend style={at={(0.53,0.95)},anchor=north west,legend cell align=left,align=left,draw=white!15!black,font=\fontsize{7}{1}\selectfont}
]
\addplot [color=blue,solid,line width=0.5pt,forget plot]
  table[row sep=crcr]{%
1	0.00574906475187212\\
2	0.00500562377340856\\
3	0.00307495133608304\\
4	0.000716334863778119\\
5	0.0012710304513862\\
6	0.00244291215218415\\
7	0.00331361784918787\\
8	0.00494093697174664\\
9	0.00756192809357444\\
10	0.01088994375687\\
11	0.0148485934606398\\
12	0.0193737744269158\\
13	0.0240261009097794\\
14	0.0279327104815123\\
15	0.0300241274145477\\
16	0.0293706329091194\\
17	0.0255756227048571\\
18	0.0192879820337352\\
19	0.01302033913165\\
20	0.011349910075191\\
21	0.0134848126764941\\
22	0.0144726226579843\\
23	0.0145268550823527\\
24	0.0153366622351775\\
25	0.011579997763479\\
26	0.0100033234387952\\
27	0.0564114314769421\\
28	0.138687112354857\\
29	0.258179390555441\\
30	0.408780914929712\\
31	0.576259132940748\\
32	0.740206678185745\\
33	0.877864586217386\\
34	0.9689143175266\\
35	1\\
36	0.967757637098733\\
37	0.87951382826266\\
38	0.75146193171577\\
39	0.604832687410109\\
40	0.461133792929248\\
41	0.337767392895785\\
42	0.245131675308779\\
43	0.185592179728403\\
44	0.153911013463299\\
45	0.139436473748349\\
46	0.13056718144226\\
47	0.118739014521291\\
48	0.0999381127218678\\
49	0.0745623945069958\\
50	0.0464101878897311\\
51	0.0221256682020853\\
52	0.0163722088071643\\
53	0.025438022512796\\
54	0.0307759795944415\\
55	0.0309142131076415\\
56	0.0277772764178651\\
57	0.0232417044310981\\
58	0.0184611331341771\\
59	0.0143152128399438\\
60	0.0117995654473675\\
61	0.0110184943623786\\
62	0.0105874305778254\\
63	0.00958031352519016\\
64	0.00896135091987709\\
65	0.0108098025165764\\
66	0.0150761252488716\\
67	0.0202121344341626\\
68	0.0250610849255791\\
69	0.0287818424124744\\
70	0.0307175654818728\\
71	0.0305249716286602\\
72	0.0282938298282325\\
73	0.0244412736054745\\
74	0.0194241944437257\\
75	0.0136063880798413\\
76	0.00780997215585861\\
77	0.00596327503401446\\
78	0.00991788433375617\\
79	0.0134478128273351\\
80	0.0142153455090663\\
81	0.0121669390569429\\
82	0.00958497779473723\\
83	0.0107236731800927\\
84	0.0144713750979972\\
85	0.016661309186989\\
86	0.0154150544854349\\
87	0.0107220664602689\\
88	0.00387134390377514\\
89	0.00317234843018982\\
90	0.00844461246904079\\
91	0.0107489878616921\\
92	0.0101261993690824\\
93	0.00837955239619383\\
94	0.00890966132800732\\
95	0.0119442732933773\\
96	0.0146844953984778\\
97	0.0157159960487469\\
98	0.014987789232471\\
99	0.0132122974322485\\
100	0.0113185036231974\\
101	0.00990527714381857\\
102	0.00905916909591588\\
103	0.00883642797111003\\
104	0.00961758981486924\\
105	0.011504541916831\\
106	0.0138499565160584\\
107	0.0157121415593553\\
108	0.0163161777921639\\
109	0.0152699553238791\\
110	0.0127291539380652\\
111	0.0097467880308507\\
112	0.00901515287544237\\
113	0.0123759764823986\\
114	0.0176519222152072\\
115	0.022845714981615\\
116	0.0268049189023447\\
117	0.0286851993083281\\
118	0.0279132801152228\\
119	0.0243297861098131\\
120	0.0182999262251037\\
121	0.0106838537787242\\
122	0.00264885640502682\\
123	0.00461722128089764\\
124	0.0101897996657322\\
125	0.0135925583060092\\
126	0.014832970710114\\
127	0.014305134978293\\
128	0.012635675451211\\
129	0.0105748741589492\\
130	0.00898616537905093\\
131	0.00870237886073431\\
132	0.00986784967430887\\
133	0.0118037889840389\\
134	0.0137426612086715\\
135	0.0151308347620463\\
136	0.0156777728007418\\
137	0.0154954466068031\\
138	0.0152058757848486\\
139	0.0156186296309638\\
140	0.0168623615332767\\
141	0.0180929517374917\\
142	0.0182553529802517\\
143	0.0167722190737708\\
144	0.0137429063693508\\
145	0.00989626143327108\\
146	0.00667019183702415\\
147	0.00634437242938157\\
148	0.00876334854004445\\
149	0.0116908691230843\\
150	0.0140355340007097\\
151	0.0152000724020061\\
152	0.0148165862917324\\
153	0.0128709732069471\\
154	0.00977503214000903\\
155	0.00636219046158981\\
156	0.00449795060249556\\
157	0.00658240955319719\\
158	0.0106484403241572\\
159	0.0148946978577958\\
160	0.0182420368176989\\
161	0.0197354670450829\\
162	0.0188078404476002\\
163	0.0155852028671506\\
164	0.0110104812580019\\
165	0.00688467962765391\\
166	0.005527377286411\\
167	0.00612360173092578\\
168	0.00596770131113042\\
169	0.00457054425774148\\
170	0.00271864413379412\\
171	0.0016951263734097\\
172	0.00191385211994984\\
173	0.00275136066187458\\
174	0.00438612231309505\\
175	0.00623775043294879\\
176	0.00734039701568274\\
177	0.0070254661619271\\
178	0.0051819034209735\\
179	0.00235907945753619\\
180	0.00178880397410922\\
181	0.00430526693047679\\
182	0.00591230330238997\\
183	0.00620146549632501\\
184	0.00539385317781832\\
185	0.00404048331294461\\
186	0.00279684263463689\\
187	0.00215377484963599\\
188	0.00213265691144841\\
189	0.00255456452648017\\
190	0.00339721217046751\\
191	0.00458431060430373\\
192	0.00600153906422159\\
193	0.00760181965560186\\
194	0.00939526664682984\\
195	0.0113861665527609\\
196	0.0135204322060175\\
197	0.0156748887352518\\
198	0.0177013718620652\\
199	0.0195202840926702\\
200	0.0212251149307373\\
201	0.0231042861296319\\
202	0.0254607866786121\\
203	0.0282676348486091\\
204	0.0309672314057389\\
205	0.0326348857269723\\
206	0.0323697246835516\\
207	0.0296849854716182\\
208	0.0247774150909498\\
209	0.0186507852851331\\
210	0.0131599950702498\\
211	0.0106136401484736\\
212	0.0110081989753616\\
213	0.0116022242445001\\
214	0.0108623016247784\\
215	0.00870018058675306\\
216	0.00554750421148513\\
217	0.00194158458267871\\
218	0.00201197774135084\\
219	0.00513670230206827\\
220	0.007183825188085\\
221	0.0078410429672259\\
222	0.007891843631788\\
223	0.00930582175869031\\
224	0.0125366947787784\\
225	0.0157203583165582\\
226	0.017056858987782\\
227	0.0157285370796291\\
228	0.0121516883729859\\
229	0.00830435687904628\\
230	0.00767473096511431\\
231	0.00972697496609361\\
232	0.0108685833181205\\
233	0.0100020765181335\\
234	0.00768110920910882\\
235	0.00528710862910943\\
236	0.00447205043419199\\
237	0.00552500899067824\\
238	0.00742462944754992\\
239	0.00943984448266848\\
240	0.0107317024920024\\
241	0.0105076000932044\\
242	0.00853953715509127\\
243	0.00552884383411698\\
244	0.00356631384239951\\
245	0.00423609623350289\\
246	0.00460386745996819\\
247	0.00366398577209684\\
248	0.00465879451405038\\
249	0.00896695707469541\\
250	0.0136246457426784\\
251	0.0169394870482539\\
252	0.0180885203083496\\
253	0.0169797299377739\\
254	0.014124910133578\\
255	0.0103877125982594\\
256	0.00680338197026463\\
};
\end{axis}
\end{tikzpicture}%
\end{minipage}
		  \begin{minipage}{0.31\linewidth}
		  \centering
%
%
\begin{tikzpicture}[scale = 1.2]
\begin{axis}[%
width=0.9in,
height=0.5in,
at={(0in,0in)},
scale only axis,
xmin=1,
xmax=256,
xlabel={Wavenum. Ind.},
xtick={1,256},
xticklabels={1,256},
xlabel style={at = {(0.5,0.35)},font=\fontsize{7}{1}\selectfont},
ticklabel style = {font=\fontsize{7}{1}\selectfont},
ymin=0,
ymax=1,
ytick={0,1},
yticklabels={0,100},
ylabel={Int.},
ylabel style={at = {(0.4,0.5)},font=\fontsize{7}{1}\selectfont},
axis background/.style={fill=white},
legend style={at={(0.53,0.95)},anchor=north west,legend cell align=left,align=left,draw=white!15!black,font=\fontsize{7}{1}\selectfont}
]
\addplot [color=red,solid,line width=0.5pt,forget plot]
  table[row sep=crcr]{%
1	0.341494613468909\\
2	0.305281601144813\\
3	0.291726702466361\\
4	0.265742844162306\\
5	0.226195707369423\\
6	0.194814819512894\\
7	0.173264183932715\\
8	0.118770684109998\\
9	0.115678632226625\\
10	0.171075829336502\\
11	0.160188637986168\\
12	0.212469137517271\\
13	0.22891322669444\\
14	0.257110425177653\\
15	0.286087462725719\\
16	0.264494668106618\\
17	0.351325437078504\\
18	0.326850391394588\\
19	0.314784338639063\\
20	0.364259953529506\\
21	0.354104855027163\\
22	0.384978313864374\\
23	0.348383420490303\\
24	0.417548534207312\\
25	0.40617464057471\\
26	0.429083446579325\\
27	0.359922038869497\\
28	0.305986282076048\\
29	0.197319136887655\\
30	0.324842990674208\\
31	0.549978749766586\\
32	0.750276425127336\\
33	0.816535254222339\\
34	0.986291989266651\\
35	1\\
36	0.989510057110804\\
37	0.867153456132697\\
38	0.768722444571133\\
39	0.596832583902979\\
40	0.454714929352358\\
41	0.343967855535357\\
42	0.162007700251209\\
43	0.0907511950413436\\
44	0.17526266845773\\
45	0.280704209198547\\
46	0.344244740946402\\
47	0.404910497259891\\
48	0.523299018249191\\
49	0.585375219318269\\
50	0.640386324658587\\
51	0.576426445704772\\
52	0.607095981649545\\
53	0.525451991021763\\
54	0.45834634054502\\
55	0.415232535727042\\
56	0.331299644396454\\
57	0.274145201437458\\
58	0.185894889188082\\
59	0.162042505066718\\
60	0.0963385701117575\\
61	0.0916799208116697\\
62	0.0820729793888043\\
63	0.0942518742746902\\
64	0.181375227307942\\
65	0.288518191518015\\
66	0.390204612287664\\
67	0.430089593645549\\
68	0.4045082919831\\
69	0.408509744988826\\
70	0.409713043700476\\
71	0.3040451381714\\
72	0.237845014162341\\
73	0.231809687850863\\
74	0.111852926228535\\
75	0.0740374933107369\\
76	0.0941080934829684\\
77	0.0987550322761422\\
78	0.108350026282936\\
79	0.141689593341619\\
80	0.171535123734325\\
81	0.130727753819153\\
82	0.0840278582288391\\
83	0.126561357766364\\
84	0.0697415129055255\\
85	0.0717484994322897\\
86	0.0766684506593714\\
87	0.0286729165619325\\
88	0.0561014001651269\\
89	0.0659550556234488\\
90	0.168621982324823\\
91	0.311490586846167\\
92	0.356692022331331\\
93	0.382701698675538\\
94	0.427226920983807\\
95	0.466895998518255\\
96	0.4720414248971\\
97	0.454730783212947\\
98	0.427327894473489\\
99	0.317099094238909\\
100	0.291052898918745\\
101	0.211096223871596\\
102	0.171798563176802\\
103	0.230724487718644\\
104	0.267002450545878\\
105	0.265616262728375\\
106	0.261560206769967\\
107	0.306865110291921\\
108	0.295281614283493\\
109	0.290452332454051\\
110	0.228765844537536\\
111	0.185736876040785\\
112	0.108154371299651\\
113	0.103477479455495\\
114	0.0998593754525426\\
115	0.23618995834788\\
116	0.314924234314388\\
117	0.377379183280849\\
118	0.360637316748298\\
119	0.402755895015273\\
120	0.429038203411143\\
121	0.372528660314031\\
122	0.39869778252023\\
123	0.340197753848622\\
124	0.346648034275401\\
125	0.309016576041926\\
126	0.220301912724884\\
127	0.166743424161079\\
128	0.0437594567473403\\
129	0.11213884105007\\
130	0.171729399270769\\
131	0.203141199697271\\
132	0.287359083688023\\
133	0.33988820674712\\
134	0.378762079345461\\
135	0.380745368843782\\
136	0.351995271994907\\
137	0.311378768691361\\
138	0.228299325695906\\
139	0.163628356794105\\
140	0.125594125893107\\
141	0.172688172865923\\
142	0.181694828372875\\
143	0.244416702503443\\
144	0.221481110180035\\
145	0.276433898856307\\
146	0.241211251786858\\
147	0.283518852620329\\
148	0.373044777791079\\
149	0.400132417001971\\
150	0.412639901115197\\
151	0.439729576474016\\
152	0.477257553037886\\
153	0.414241930675734\\
154	0.424500589885213\\
155	0.394285400941149\\
156	0.393004697203603\\
157	0.363616285830166\\
158	0.360530656867844\\
159	0.246364542743581\\
160	0.188654488108006\\
161	0.198186067648745\\
162	0.197515644735294\\
163	0.180703899304927\\
164	0.222012043616512\\
165	0.24815958421333\\
166	0.242164402222587\\
167	0.238304006691012\\
168	0.243980632403218\\
169	0.187180481261699\\
170	0.162371885367106\\
171	0.174366108048944\\
172	0.127977444594647\\
173	0.0878170501354195\\
174	0.0554956095887277\\
175	0.0544818579329285\\
176	0.098234968383092\\
177	0.0778409416932297\\
178	0.0662309907263098\\
179	0.0935821039660997\\
180	0.174912526931355\\
181	0.214443114380935\\
182	0.204970943263423\\
183	0.234778668398475\\
184	0.211086851552661\\
185	0.242879769205948\\
186	0.173228820171938\\
187	0.202534411115812\\
188	0.200822990520062\\
189	0.242718036717426\\
190	0.236246427681634\\
191	0.229156725709221\\
192	0.262592420643492\\
193	0.203441035720229\\
194	0.252496972973522\\
195	0.273766851371261\\
196	0.230628390316368\\
197	0.213251186395836\\
198	0.276625984485799\\
199	0.278974968719304\\
200	0.231750955200057\\
201	0.314858077492785\\
202	0.237158492317908\\
203	0.207155725062241\\
204	0.196773255039441\\
205	0.0863502073785862\\
206	0.0902953649299595\\
207	0.0868356229541647\\
208	0.107425186108479\\
209	0.156421225664395\\
210	0.0995250456022564\\
211	0.0708876728152659\\
212	0.0793206416511559\\
213	0.156118492699843\\
214	0.271528416068609\\
215	0.371145002593212\\
216	0.411674806400622\\
217	0.507756672549762\\
218	0.528183944661258\\
219	0.441167865047642\\
220	0.422537615016984\\
221	0.35328468964781\\
222	0.292941735997408\\
223	0.248471985582846\\
224	0.301349374131025\\
225	0.374599539178608\\
226	0.421467775363944\\
227	0.479736322024826\\
228	0.484818423482492\\
229	0.444702289021447\\
230	0.494323562480683\\
231	0.410075969140507\\
232	0.429076258154407\\
233	0.377100787266215\\
234	0.370580467523857\\
235	0.306343398511173\\
236	0.365183191195551\\
237	0.348159575797996\\
238	0.363386726780381\\
239	0.387855671281607\\
240	0.338377001002682\\
241	0.364703123551825\\
242	0.274045379646807\\
243	0.288414286268181\\
244	0.242361387559345\\
245	0.218623550332459\\
246	0.167662408893999\\
247	0.124789419476966\\
248	0.150197182660384\\
249	0.138451936340921\\
250	0.188374039085216\\
251	0.231026515755355\\
252	0.273039040080369\\
253	0.286022663954863\\
254	0.268153544930923\\
255	0.301941090637059\\
256	0.305670222221405\\
};
\end{axis}
\end{tikzpicture}%
		  \end{minipage}
\end{minipage}
\caption{The reconstructed HS volumes. (top) The spatial maps at 594 nm. (bottom) The spectra at the center pixel. The value of Signal-to-Reconstruction Error (SRE) $:=10 \log (\|\bs X_\textsl{ref}\|^2_F/\| \bs X_\textsl{ref} - \hat{\bs X}\|^2_F)$ indicates the superiority of the proposed method.}
\label{fig:result}
\end{figure}

We simulate the SP-FTI model defined in \eqref{eq:SPFTI acquisition model} from actual FTI measurements recorded at Nyquist regime (see \cite{moshtaghpour2018} for a detailed description of the related experiment). In summary, a thin layer of a transparent cell, \ie Convallaria, lily of the valley is observed through the FTI procedure. From the Nyquist FTI measurements of size $(N_\xi,N_x,N_y) = (512,128,128)$ we then form SP-FTI measurements of size $(M_\xi,M_p) = (112,8218)$, \ie equivalent to MUR = ERR = $0.11$. In Fig.~\ref{fig:result}, we compare a recovered instance of our approach with \emph{(i)} a reference HS volume $\bs X_\textsl{ref}$, recovered from Nyquist observations, and \emph{(ii)} the HS volume recovered from SP-FTI measurements with UDS strategy, all volumes being recovered via \eqref{eq:recovery}. As can be seen, SP-FTI combined with MLS strategy yields high quality HS volume, as opposed to the UDS strategy.

\section{Conclusion}
We presented a proof of concept for a new compressive Hyperspectral acquisition framework. Our approach includes \emph{(i)} space-time illumination coding, and \emph{(ii)} single-pixel imaging techniques. In SP-FTI we adapted the best illumination/aperture coding strategy, using multilevel sampling principle, suitable for FS. We obtained successful theoretical and numerical results in terms of light exposure reduction and low undersampling ratio. The latter, however, motivates us to consider the application of portable SP-FTI in airborne systems as our future work. 
\bibliographystyle{IEEETran}
\bibliography{IEEEabrv,refs}
\end{document}